\documentclass[conference]{IEEEtran}
\usepackage{cite}
\usepackage{amsmath,amssymb,amsfonts}
\usepackage{algorithmic}
\usepackage{graphicx}
\usepackage{textcomp}
\usepackage{xcolor}

\usepackage{booktabs} 
\usepackage{multirow}
\usepackage{subcaption}
\usepackage{siunitx}
\usepackage{color}
\usepackage{soul}
\usepackage{algorithm}
\definecolor{tbc}{rgb}{1.0,0,0}

\def\BibTeX{{\rm B\kern-.05em{\sc i\kern-.025em b}\kern-.08em
    T\kern-.1667em\lower.7ex\hbox{E}\kern-.125emX}}
\begin{document}

\title{DOPING: Generative Data Augmentation for Unsupervised Anomaly Detection with GAN}

\author{\IEEEauthorblockN{Swee Kiat Lim, Yi Loo, Ngoc-Trung Tran, Ngai-Man Cheung, Gemma Roig, Yuval Elovici}
\IEEEauthorblockA{\textit{ST Electronics - SUTD Cyber Security Laboratory, Singapore University of Technology and Design} \\
Singapore \\
\{sweekiat\_lim, loo\_yi, ngoctrung\_tran, ngaiman\_cheung, gemma\_roig, yuval\_elovici\}@sutd.edu.sg
}}

\maketitle

\begin{abstract}
Recently, the introduction of the generative adversarial network (GAN) and its variants has enabled the generation of realistic synthetic samples, which has been used for enlarging training sets. 
Previous work primarily focused on data augmentation for semi-supervised and supervised tasks.
In this paper, we instead focus on unsupervised anomaly detection and propose a novel generative data augmentation framework optimized for this task.
In particular, we propose to oversample {\em infrequent normal samples} - normal samples that occur with small probability, e.g., rare normal events.
We show that these samples are responsible for false positives in anomaly detection. 
However,
oversampling of infrequent normal samples is challenging for real-world high-dimensional data with  multimodal distributions. To address this challenge, we propose to use 
 a GAN variant known as the adversarial autoencoder (AAE)
to transform the high-dimensional multimodal data distributions into low-dimensional unimodal latent distributions with well-defined tail probability. Then, we systematically oversample at the ``edge'' of the latent distributions to increase the density of infrequent normal samples.  
We show that our oversampling pipeline is a  
unified one: it is  generally applicable to  datasets with different complex data distributions.
To the best of our knowledge, our method is the first data augmentation technique focused on improving performance in unsupervised anomaly detection. We validate our method by demonstrating consistent improvements across several real-world datasets\footnote{\scriptsize Corresponding author: Ngai-Man Cheung (ngaiman\_cheung@sutd.edu.sg).
Earlier version was accepted in  
 IEEE International Conference on Data Mining 2018 (ICDM-18).
Our code will be published here: https://github.com/greentfrapp/doping.
}.
\end{abstract}


\begin{figure}
\begin{center}
\includegraphics[width=9cm]{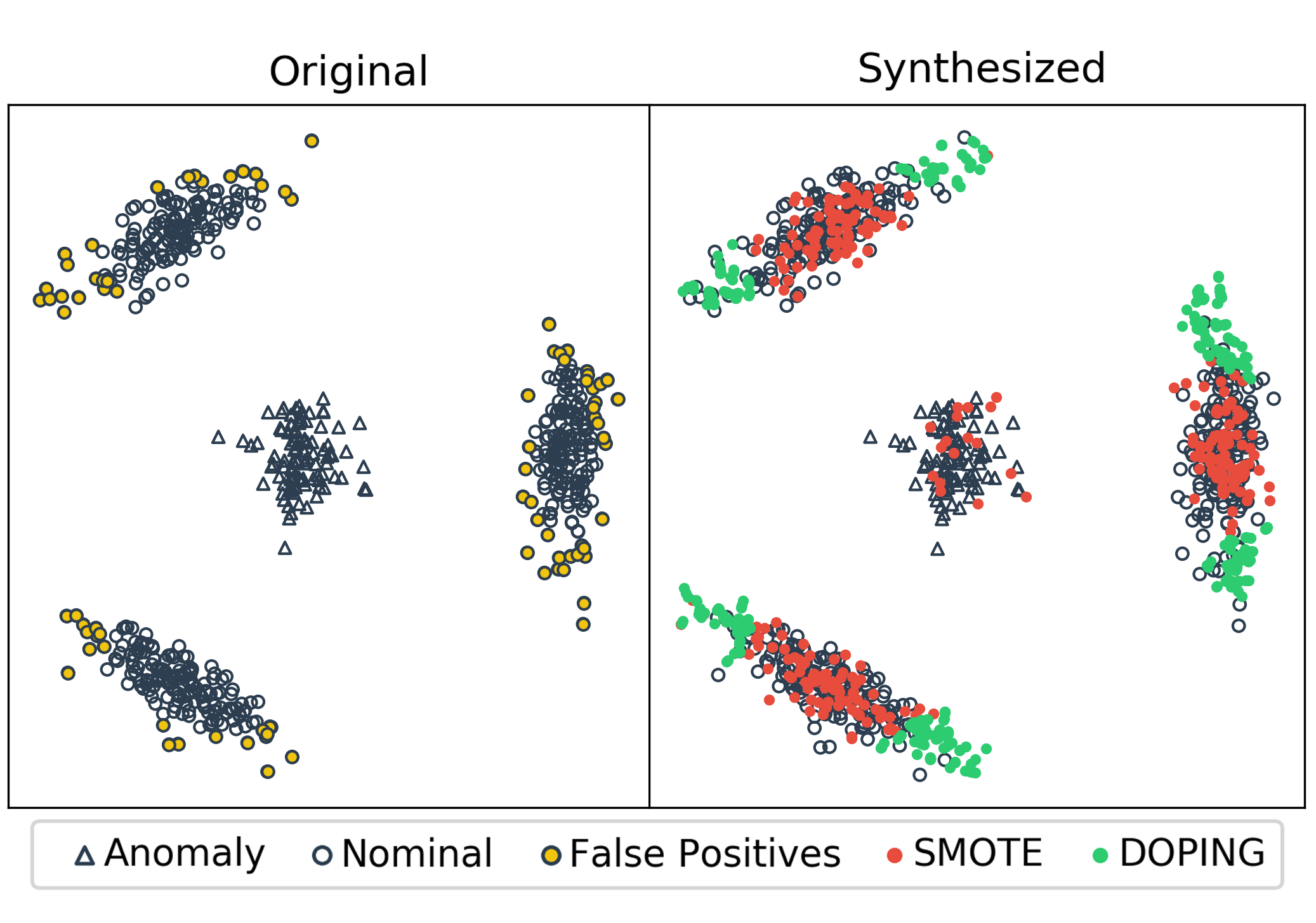}
\end{center}
\caption{Our DOPING method is a form of unsupervised data augmentation. Consider the above 2D dataset with 5\% contamination, where the anomalies are clustered in the middle.
Note that we focus on unsupervised setting, and the class labels are unknown to the methods. In this case, 
it is unclear how data augmentation should be performed since no class label is provided. Using an interpolation method such as in SMOTE \cite{chawla2002}, we can randomly synthesize samples across the training set (red), but this increases the density of all samples indiscriminately, which is not helpful for most unsupervised anomaly detection algorithms. Instead, we propose DOPING, a method that strategically synthesizes samples at the edge of the normal distribution (green). Our DOPING samples help to reduce false positive rates by increasing the density of 
infrequent normal samples.
In the above example, both the False Positives and DOPING samples occur in overlapping areas.
Using our novel idea of sampling in the latent distribution imposed by an adversarial autoencoder (AAE),
infrequent normal samples can be easily generated even for high-dimensional data with multi-modal distributions. 
We emphasize that we achieve this without using class labels.
}
\label{fig:frontispiece}
\vspace{-0.5cm}
\end{figure}

\section{Introduction}
Data augmentation and oversampling are important approaches to handle imbalanced data in machine learning.
Previous data augmentation and oversampling approaches have focused on the supervised setting. For example, 
Synthetic  Minority Over-sampling Technique (SMOTE) \cite{chawla2002} and Structure Preserving Oversampling (SPO) \cite{cao2011} 
use minority class samples to generate synthetic samples.
In such methods, class labels are essential for identifying and selecting these minority samples.

In this work, instead, we focus on the unsupervised setting. This is important for many anomaly detection problems: the anomalous events, which are the minority samples, may be undetected or unknown previously, e.g. a zero-day cyber-attack.
For unsupervised anomaly detection, previous data argumentation approaches are ineffective, since labels and knowledge of minority classes are unavailable to the systems (Figure \ref{fig:frontispiece}). 

One of the key challenges of anomaly detection systems is the pervasiveness of false positives~\cite{chandola2009anomaly} (positives are anomalous events). This is especially true when high recall is required, such as in malware detection, where it is preferable to raise a false alarm rather than let slip an undetected virus.  
High false positive rates can be due to the difficulty in defining a distribution which contains all of the normal data. 
In particular, {\em infrequent normal samples}, defined here as normal samples that occur with very small probability, occur rarely in the training dataset and are poorly characterized in many density-based anomaly detectors \cite{breunig2000, liu2008}. 
The boundary between normal and anomalous data is typically imprecise, resulting in misclassification of samples close to the boundary \cite{chandola2009anomaly}.

We hypothesize that in unsupervised anomaly detection, targeted oversampling of infrequent normal samples will be most effective, since such samples are primarily responsible for false positive errors. In this paper, we first proceed to validate this hypothesis in our study. Based on this idea, we then introduce a novel data augmentation method for systematically generating such infrequent normal samples without using class labels.

Specifically,  we propose a data augmentation method targeted at unsupervised anomaly detection, which we coin DOPING. 
Generating infrequent normal samples can be difficult for high-dimensional data with complicated distributions,
e.g., multi-modal
(Figure \ref{fig:frontispiece}). 
We propose to overcome this challenge by 
using the adversarial autoencoder (AAE) \cite{makhzani2015}, a variant of the generative adversarial network (GAN) \cite{goodfellow2014}.
In particular, using the AAE, we impose a multivariate Gaussian on the latent space of the training data, thereby transforming different data distributions into a unimodal distribution with well-defined tail probability in the latent space.  Then, infrequent normal samples can be easily generated by sampling this unimodal latent distribution.  
We will go on to show that our unified and systematic latent space sampling can handle diverse datasets with complicated data distributions.
In summary, our DOPING technique differs both from previous works where GAN architectures were used in anomaly detection \cite{schlegl-ipmi-2017, ravanbakhsh-arxiv-2017, zenati2018}, as well as previous works on data augmentation for supervised classification \cite{perez2017effectiveness, antoniou2017, zhu2017, sixt2016, wong2016, simard2003}:
\begin{itemize}
	\item It is an unsupervised technique that does not require any normal or anomalous labels. Our proposed method trains the AAE on the entire dataset without labels and utilizes general knowledge of the latent distribution to generate desired samples. 
	\item Rather than an independent anomaly detection technique, DOPING is a form of unsupervised data augmentation. This makes it complimentary with any anomaly detection algorithm, by augmenting the dataset before training. 
	\item DOPING differs from previous data augmentation methods, in that it uses a systematic sampling strategy that has been validated in our analysis to be effective for anomaly detection. In contrast, previous methods mainly focused on generating artificial samples in a minority class.
\end{itemize}

In the following sections, we discuss details of our DOPING method for unsupervised anomaly detection. We perform studies to validate the effectiveness of oversampling infrequent normal samples. Then, we compare DOPING to other data augmentation techniques and validate our method on several real-world datasets.

\section{Related Work}



Autoencoders have previously been used directly for anomaly detection \cite{munawar-arxiv-2017, hasan-cvpr-2016, zhou2017, seebock-arxiv-2016, xu2015learning}. Munawar \emph{et al.} \cite{munawar-arxiv-2017} used both normal and anomalous labeled data to train an autoencoder by minimizing reconstruction loss of normal data and maximizing the loss for anomalous data. The trained autoencoder then detects anomalies by attempting to reconstruct test samples and measuring reconstruction loss, which is larger for anomalies. 
A recent work by Zhou \emph{et al.} \cite{zhou2017} also introduced robust deep autoencoders (RDA), an unsupervised anomaly detection method that combined the autoencoder and Robust Principal Component Analysis (PCA). The model is trained to decompose the dataset as the sum of two matrices - a normal component that can be reconstructed with little loss via an autoencoder and a sparse matrix consisting of anomalies.


Recent anomaly detection methods have also utilized the GAN \cite{schlegl-ipmi-2017, ravanbakhsh-arxiv-2017, zenati2018}. Typically, such methods are similar to the approach above where autoencoders are used directly for anomaly detection and are trained on a known normal dataset. Schlegl \emph{et al.} \cite{schlegl-ipmi-2017} trained a GAN on the normal dataset and defined a loss function for mapping a new sample to a latent vector. The anomaly score is given as the difference between the sample generated by the latent vector and the original sample. Similarly, Ravanbakhsh \emph{et al.} \cite{ravanbakhsh-arxiv-2017} trained a conditional GAN \cite{isola-cvpr-2017} to learn two generators that map pixel data to motion and vice versa. Normal frames will then have low reconstruction loss, whereas anomalous frames will be poorly reconstructed. More recently, Zenati \emph{et al.} \cite{zenati2018} utilized the BiGAN \cite{donahue2016} for anomaly detection by training on normal data and then defining a score function based on reconstruction loss and discriminator-based loss.


As a generative model, GANs have also been used for data augmentation, generating samples in imbalanced or small datasets, particularly in supervised classification \cite{perez2017effectiveness, antoniou2017, zhu2017, sixt2016}. More traditional data augmentation techniques for synthesizing data samples include geometric transformation or oversampling \cite{wong2016, simard2003}.

Although our work uses the AAE, which combines the GAN and the autoencoder, it is fundamentally different from previous approaches, since ours is an unsupervised approach and does not require a known set of normal samples or labels. Furthermore, our novel method is a form of unsupervised data augmentation that can be applied to any dataset or anomaly detection algorithm. 

Several oversampling techniques have been proposed: SMOTE \cite{chawla2002}, Borderline-SMOTE \cite{han2005} and SPO \cite{cao2011} and Integrated Oversampling (INOS) \cite{cao2013}. However, these techniques focus on oversampling the minority class for use in supervised classification. In contrast, our method focuses on unsupervised anomaly detection, where we do not have access to label information  of the minority samples, which is required for the aforementioned methods. Our data augmentation scheme is derived from our analysis and specifically targets unsupervised anomaly detection. In particular, we show that conventional data augmentations do not lead to optimal improvement in unsupervised anomaly detection.

In this paper, we propose a new mechanism to generate synthetic data for improving anomaly detection systems in a purely unsupervised setting. To the best of our knowledge, our method is the first data augmentation technique focused on improving performance and reducing false positive rates in unsupervised anomaly detection. We also compare against previous data augmentation methods and show more consistent and larger improvements in performance.


\begin{figure*}
	\centering
	\includegraphics[width=17cm]{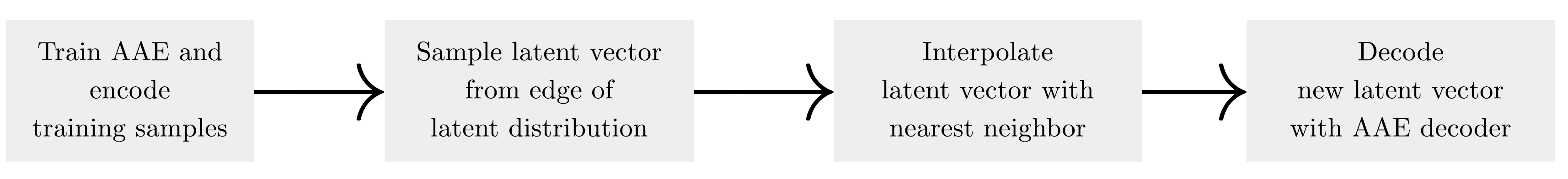}
	\caption{Data augmentation with DOPING: System overview}
    \label{fig:schematic}
\vspace{-0.3cm}
\end{figure*}

\section{DOPING}
\label{sec:doping}

We hereby introduce Doping with Infrequent Normal Generator (DOPING), a data augmentation technique for anomaly detection algorithms (Figure \ref{fig:schematic}). In physics, doping is a process whereby a material is introduced into a semiconductor, typically to improve its electrical conductivity. Similarly, our method introduces artificial samples to the original training set, to improve the performance of anomaly detection algorithms. These artificial samples take the form of infrequent normal samples that are synthetically generated with AAEs.

\noindent \textbf{Adversarial Autoencoders (AAE)~\cite{makhzani2015}.} Consider a dataset with data distribution $p_d(\mathbf{x})$ and an encoding function $q(\mathbf{z}|\mathbf{x})$ from the autoencoder. The aggregated posterior distribution $q(\mathbf{z})$ on the latent vector is then defined as:
\begin{equation*}
q(\mathbf{z}) = \int_\mathbf{x}q(\mathbf{z}|\mathbf{x})p_d(\mathbf{x})\mathrm{d}\mathbf{x}
\end{equation*}
The AAE tries to match $q(\mathbf{z})$ to an arbitrary prior $p(\mathbf{z})$, similar to the variational autoencoder (VAE) \cite{kingma2013}. While the VAE uses the KL-divergence for regularization, the AAE borrows the adversarial concept from the GAN and attaches a discriminator network that discriminates between outputs from the prior $p(\mathbf{z})$ and the encoder $q(\mathbf{z})$. After training, $q(\mathbf{z})$ then matches $p(\mathbf{z})$. Using this mechanism, \cite{makhzani2015} demonstrated several interesting latent spaces including a mixture of 10 2-D Gaussians and a swiss roll distribution.

\subsection{Data augmentation with DOPING: Details}
\label{sec:implementation}

\noindent \textbf{Train Unlabeled AAE} In DOPING, we apply the vanilla Unlabeled AAE architecture \cite{makhzani2015}. The encoder, decoder and discriminator each have two layers of 1000 hidden units with ReLU activation function, except for the activation of the last layer of the decoder, which is linear. {\em No label is used in the training of the Unlabeled AAE}. We use the ADAM optimizer \cite{kingma2014} with a learning rate of $10^{-4}$ for all networks. We use a 2-D Gaussian with mean 0.0 and standard deviation 10.0 on both dimensions as the prior $p(z)$.
Note that other recent GAN with encoder and decoder can also be used \cite{trung:2018:distgan}.

\noindent \textbf{Sample Latent Space} After training the AAE, we selectively sample and decode latent vectors to generate synthetic samples. As will be discussed, our analysis finds improved performance when we decode latent vectors at the boundary of the normal latent distribution. Hence, we formalize the DOPING technique as in Algorithm \ref{alg:doping} and use an edge-based sampling approach to sample the latent vectors for decoding. 

Specifically, we use the AAE's encoder $E(\cdot)$ to encode the entire training set $X$ and generate the corresponding set of latent vectors $Z$. $Z$ is then filtered by the norm of the latent vectors to form the subset $Z_\text{edge}$ (see Equation \ref{eqn:edge}), which contains latent vectors near the tail-end of the latent distribution.


\begin{equation}
	Z_\text{edge} = \{\mathbf{z}\in Z \mid \alpha<\left\|\mathbf{z}\right\|<\beta\}
\label{eqn:edge}
\end{equation}

Based on a validation dataset, we fix $\beta$
and $\alpha$ as follows for  all experiments in this paper: we set $\beta$ as 3 standard deviations larger than the mean of the latent vector norms for the training set; we set $\alpha$ as the 90th percentile norm from the remaining vectors.

\noindent \textbf{Interpolate Sampled Vectors} Thereafter, we randomly sample latent vectors from $Z_\text{edge}$ and new latent vectors are generated by interpolating these selected latent vectors with their nearest neighbor in the latent space (Algorithm \ref{alg:internn}). These new latent vectors form the set $Z_\text{synth}$.

\noindent \textbf{Decode and Synthesize Samples} We then decode the set of new latent vectors $Z_\text{synth}$ to generate synthetic samples $X_\text{synth}$ for data augmentation. The synthetic samples $X_\text{synth}$ are added to the original dataset $X$ and the augmented dataset is used to train the anomaly detector algorithm.

\begin{algorithm}
\caption{InterNN (interpolate with nearest neighbor)}
\label{alg:internn}
\begin{algorithmic}
\REQUIRE $\mathbf{z}_{\text{sample}}$: latent vector to interpolate
\STATE Sample $\alpha \sim U(0, 1)$
\STATE $\mathbf{z}_{\text{NN}} \leftarrow$ nearest neighbor of $\mathbf{z}_{\text{sample}}$ in latent space
\RETURN $\alpha(\mathbf{z}_{\text{NN}} - \mathbf{z}_{\text{sample}}) + \mathbf{z}_{\text{sample}}$
\end{algorithmic}
\end{algorithm}


\begin{algorithm}
\caption{DOPING}
\label{alg:doping}
\begin{algorithmic}
\REQUIRE $K$: no. of samples to synthesize, $X$: set of training samples, $\{\alpha,\beta\}$: hyperparameters for edge-based sampling (see Equation \ref{eqn:edge})
\STATE Train AAE on $X$, with multivariate Gaussian as prior
\STATE $E(\cdot) \leftarrow$ encoder in AAE
\STATE $D(\cdot) \leftarrow$ decoder in AAE
\STATE $Z \leftarrow E(X)$
\STATE $Z_\text{edge} \leftarrow \{\mathbf{z}\in Z \mid \alpha<\left\|\mathbf{z}\right\|<\beta\}$ (see Section \ref{sec:implementation})
\STATE Initialize list of new latent vectors $Z_{\text{synth}} \leftarrow \text{[]}$
\FOR{$i = 1, 2, ... K$}
\STATE Sample infrequent latent vector $\mathbf{z}_{\text{edge}} \in Z_{\text{edge}}$
\STATE $\mathbf{z}_{\text{new}} \leftarrow \text{\textbf{InterNN}}(\mathbf{z}_{\text{edge}})$
\STATE $Z_{\text{synth}}$ \textbf{append} $\mathbf{z}_\text{new}$
\ENDFOR
\STATE $X_\text{synth} \leftarrow D(Z_\text{synth})$
\STATE Train anomaly detector on augmented dataset $(X+X_\text{synth})$
\end{algorithmic}
\end{algorithm}

\subsection{Use of Adversarial Autoencoders}
We specifically choose to adopt AAE over other GAN variants in our framework for the following reasons:

\noindent \textbf{Explicit Encoding to Latent Space} While other GAN variants are able to synthesize new samples from a latent space after training, it is non-trivial to encode an arbitrary sample back into the latent space, e.g., requiring iterative estimation \cite{schlegl-ipmi-2017}.
In contrast, the AAE incorporates an autoencoder architecture, which allows explicit decoding from and encoding to the latent space. The encoding network is crucial for our analysis of samples in the latent space, while the decoding network enables generation of synthetic samples from specific regions in the latent space based on our analysis. The encoding and decoding are tightly coupled in AAE, enabling our analysis outcomes to directly guide sample synthesis.  

\begin{figure*}[h!]
	\begin{center}
	\begin{subfigure}[b]{13cm}
	\centering
	\includegraphics[width=13cm]{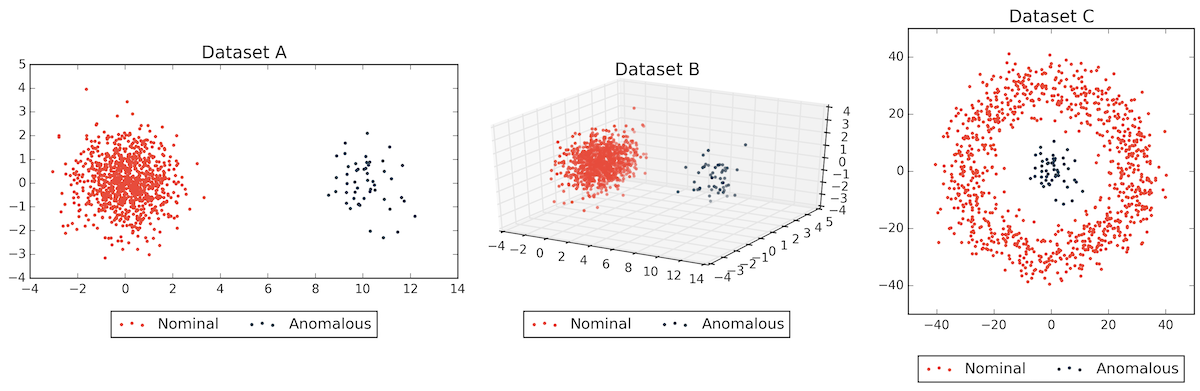}
	\caption{}
    \label{fig:toy_expt_datasets}
    \end{subfigure}
    \begin{subfigure}[b]{16cm}
	\centering
	\includegraphics[width=16cm]{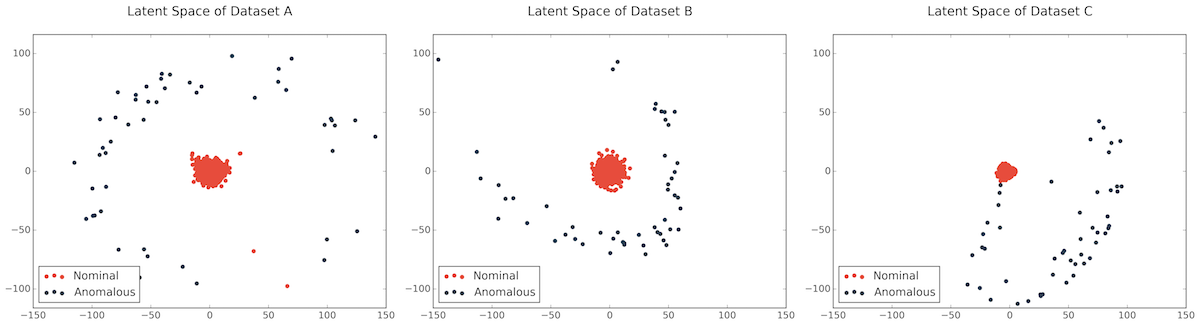}
	\caption{}
    \label{fig:toy_expt_latent_spaces}
	\end{subfigure}
    \end{center}
\caption{(a) The three synthetic datasets used and (b) the corresponding latent spaces after training with the AAE.}
\vspace{-0.5cm}
\end{figure*}

\noindent \textbf{Explicit Control of Latent Space} Using the AAE, we are able to  impose a variety of prior distributions on the latent space, similar to the VAE \cite{kingma2013}. However, the VAE requires access to the exact functional form of the prior distribution. In contrast, the adversarial method used by AAE only requires samples from the desired prior, which allows us to impose more complex distributions.
By imposing the same prior distribution and systematically decoding from this consistent latent space, we have a general method that can be applied to any dataset.

In our method, we use the AAE to impose a multivariate Gaussian on the latent space of the normal data and treat that as a proxy for the data distribution. 
Our proposed use of multivariate Gaussian as the prior transforms different complex data distributions (e.g. multimodal, skewed) into {\em unimodal ones with well-defined tail probability in the latent space}.
This enables decoding from the edge of the latent Gaussian distribution to generate synthetic samples that are infrequent normals.
As a data augmentation technique, DOPING is used with another anomaly detection algorithm, such as Isolation Forest (iForest) \cite{liu2008}. 


\section{Analysis of DOPING with Synthetic Datasets}
\label{sec:toy_expt}

In this section, we analyze and validate fundamental components of DOPING  with synthetic datasets. We first demonstrate that the use of the AAE in our framework enables mapping of  diverse data distributions to a consistent latent distribution. This enables a unified analysis, which reveals that there exists a consistent region in the latent space for generating synthetic samples that improves the anomaly detection performance. Most importantly, this sampling method for data augmentation is consistent despite the different data distributions.

\subsection{Dataset and Experimental Design}
\label{sec:toy_expt_setup}

\noindent \textbf{Dataset} We introduce three synthetic datasets that each comprise samples from a normal distribution and an anomalous distribution
(Figure \ref{fig:toy_expt_datasets}).
Each dataset contains 1000 training samples and 1000 test samples, with 95\% of the samples belonging to the normal distribution and 5\% of the samples belonging to the anomalous distribution.

\begin{itemize}
	\item \textbf{Dataset A} The normal distribution is a 2-D Gaussian with mean at origin and standard deviation 10.0 on each dimension, while the anomalous distribution is a 2-D Gaussian with mean [30.0, 0.0] and standard deviation 5.0 on each dimension.
	\item \textbf{Dataset B} A 3-D version of Dataset A. 
	\item \textbf{Dataset C} The normal distribution follows a ring with radius of mean 30.0 and standard deviation 5.0, while the anomalous distribution is a 2-D Gaussian with mean at origin and standard deviation 5.0 on each dimension. This is meant to be a challenging dataset, with the anomalies surrounded by normal samples, making it more difficult for anomaly detection algorithms to recognize anomalies.
\end{itemize}

\noindent \textbf{Baseline Method} In this experiment, we use the popular iForest anomaly detection algorithm. We use the implementation in version 0.19.1 of the scikit-learn package \cite{scikit-learn} and vary the contamination hyperparameter from 0.01 to 0.69 in increments of 0.04 to generate the receiver operating characteristic (ROC) curve for evaluation.

\noindent \textbf{Experiment Details} In this section, we first adopt a {\em Labeled AAE} architecture to analyze our method. It is similar to the Unlabeled AAE, with the exception of label information provided to the discriminator network in the form of a one-hot vector. In our case, the label indicates if the sample is normal or anomalous. If the sample is anomalous, we use a ring with radius 100.0 as the prior $p(z)$, by randomly sampling a latent vector with $l^2$-norm of 100.0. If the sample is normal, we follow the prior used in the Unlabeled AAE. All other conditions are the same as described in the Unlabeled AAE.
This Labeled AAE is used {\em only in this analysis} to understand the effectiveness of sampling at different latent regions.  In practice, Unlabeled AAE is used, as will be further discussed.

In this section, we also use a magnitude-based sampling method for synthesizing new samples, in order to demonstrate the effects of sampling from different regions in the latent distribution. In magnitude-based sampling, we sample the latent vectors from an n-dimensional spherical surface of increasing magnitudes. In this case, with a 2-D latent space, we essentially sample from rings of increasing magnitudes.

\begin{figure*}
\includegraphics[width=18cm]{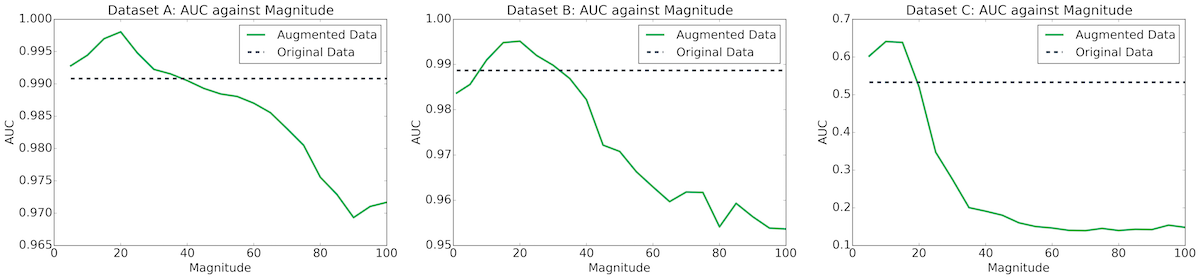}
\caption{Graphs of AUC against $l^2$-norm/magnitude for the different synthetic datasets, showing a consistent trend.} 
\label{fig:toy_expt_auc}
\vspace{-0.5cm}
\end{figure*}

We sample latent vectors at different $l^2$-norms from 5.0 to 100.0 at increments of 5.0. At each $l^2$-norm, we decode and add the 100 random samples to the original dataset. Each augmented dataset of 1100 samples is then used to train the anomaly detectors and tested on the test set, to generate the ROC curves. We then measure the area-under-curve (AUC) for each ROC curve and plot the AUC against the $l^2$-norm, as shown in Figure \ref{fig:toy_expt_auc}. We treat predictions of anomalies as positive instances for the calculations of the AUC.


\begin{figure}
	\begin{center}
	\begin{subfigure}[b]{6cm}
	\centering
	\includegraphics[width=6cm]{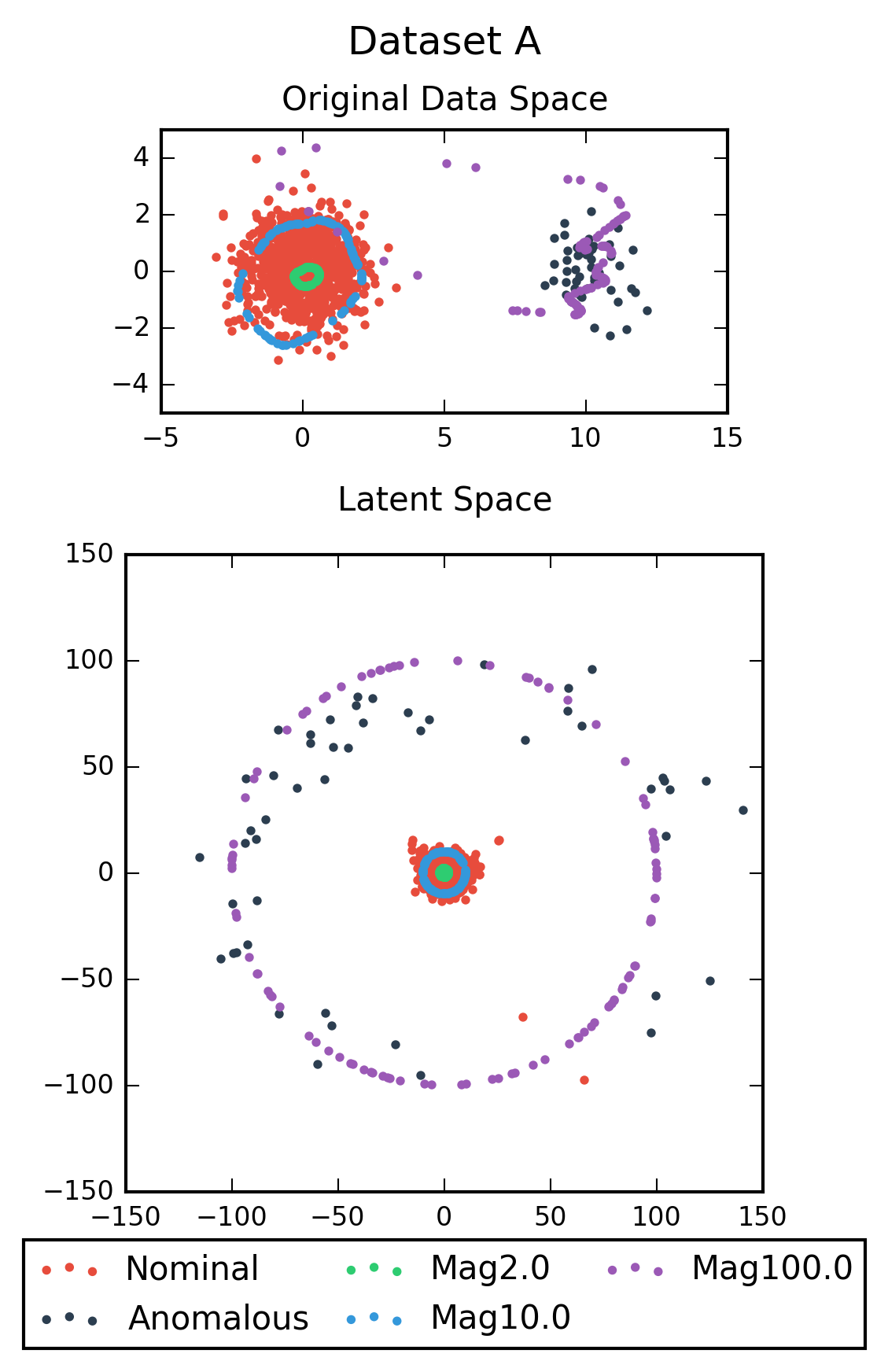}
	\caption{}
	\end{subfigure}
    \begin{subfigure}[b]{6cm}
	\centering
	\includegraphics[width=6cm]{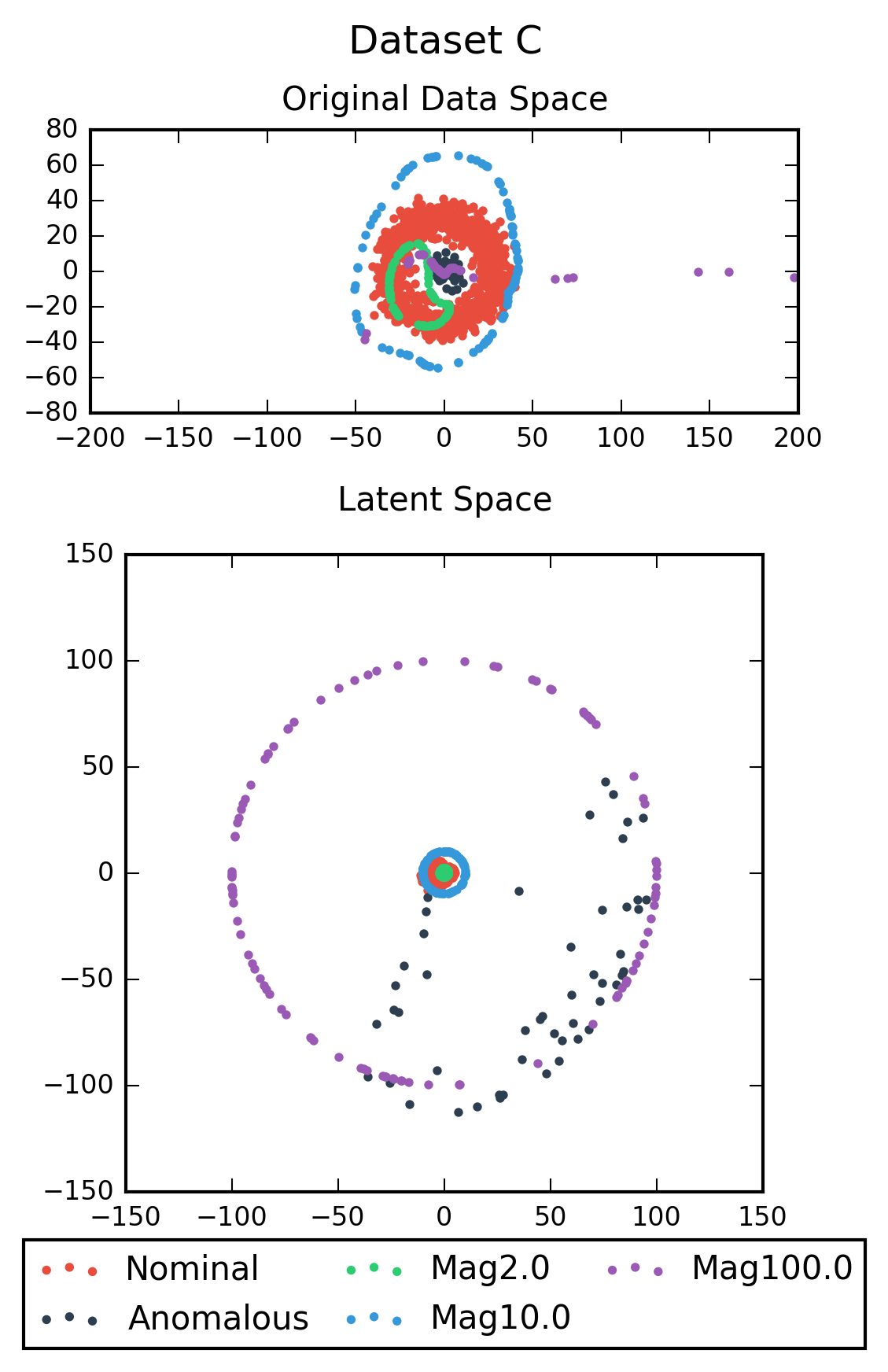}
	\caption{}
	\end{subfigure}
    \end{center}
\caption{Visualization of samples decoded from the latent space of (a) Dataset A and (b) Dataset C. Latent vectors at the boundary of the latent normal distribution (blue) decode to points at the boundary of the normal distribution in the original data space.}
\label{fig:toy_expt}
\end{figure}

\subsection{Mapping of Data Spaces to Latent Distributions}
\label{sec:mapping}

Figure \ref{fig:toy_expt_datasets} shows 3 different datasets. By training an AAE with the same prior on each of these datasets, we are able to map these datasets to similar latent distributions, in Figure \ref{fig:toy_expt_latent_spaces}. In particular, this happens even for Dataset C, which is particularly challenging since the AAE has to switch the distributions of the normal and anomalous samples.

Furthermore, we are able to systematically sample the latent space to generate corresponding samples in the data space. 
With the use of 2-D Gaussian, which has well-defined tail probability, as the prior, 
we can sample latent vectors of different $l^2$-norms, corresponding to different probabilities or likelihoods. This is akin to sampling latent vectors at rings of different radii. Since the anomalous latent distribution forms a ring around the normal distribution, as we increase the $l^2$-norm of the sampled latent vectors, the generated samples should intuitively transform from normal to anomalous. Figure \ref{fig:toy_expt} shows the data samples generated by decoding from different rings in the latent space for Datasets A and C. 


\subsection{Optimal Latent Space Sampling for Synthesizing Samples}
\label{sec:optimalregion}

In this section, we test our hypothesis that decoding from the boundary of the normal latent distribution to generate infrequent normal samples provides improvement in anomaly detection performance, when such samples are added to the training set.

All three graphs in Figure \ref{fig:toy_expt_auc} clearly show the same trends, with significantly better AUC performance when DOPING with latent vectors of $l^2$-norms in the range of 15.0 to 20.0. With reference to Figure \ref{fig:toy_expt_latent_spaces}, this region corresponds to the boundary of the normal latent distribution, which agrees with our hypothesis.

The optimal samples for data augmentation is not necessarily intuitive when viewed in the data space. For instance, in Dataset C, it is not immediately clear if the ideal synthetic samples should lie in the inner or outer boundary of the normal distribution. Despite that, Figure \ref{fig:toy_expt_auc} shows that an improvement can be consistently achieved with our method, when decoding from the boundary of the latent distribution. 
Referencing to Figure \ref{fig:toy_expt}  reveals that latent vectors at the boundary of the normal distribution map to data samples on the outer boundary of the normal data. 

Thereafter, as $l^2$-norm increases, the AUC falls to below that of the original dataset, implying that addition of samples in this region actually worsens the performance of the anomaly detectors. In the context of our hypothesis, samples decoded from latent vectors in this region corresponds to anomalous samples. By increasing the density of anomalies in the training set, the anomaly detector algorithms are more likely to misclassify anomalies as normal, which worsens performance.

\textbf{Unsupervised Data Augmentation} Since samples at the boundary of the normal latent distribution provide improvement in anomaly detection performance, in practice, the Labeled AAE can be replaced with the unsupervised Unlabeled AAE. We reiterate that the nature of anomaly detection tasks is that ``anomalies are `few and different', which make them more susceptible to isolation than normal points" \cite{liu2008}. With low contamination, an Unlabeled AAE trained on the entire dataset can approximate the normal latent distribution in a Labeled AAE and the contamination due to the 'few and different' anomalies can be neglected. We demonstrate this in the following sections where we only use the Unlabeled AAE.

\section{Analysis of DOPING with MNIST Experiment}
\label{sec:mnist_expt}

In the previous section we introduced a consistent data augmentation method to improve anomaly detection performance. Here we validate this methodology on the MNIST dataset, a more complex dataset with much higher dimension, and show that the method works in an unsupervised manner with the Unlabeled AAE. We also compare different prior distributions for the Unlabeled AAE. We then compare our method against other data augmentation methods across several state-of-the-art anomaly detectors and show that our method gives consistent and substantial improvement.

\subsection{Dataset and Experimental Setup}
\label{sec:mnist_setup}

\noindent \textbf{Dataset} We follow the setup used by Zhou \emph{et al.} in their paper on anomaly detection with Robust Deep Autoencoders (RDA) \cite{zhou2017}. Our dataset comprises 5000 MNIST \cite{lecun1998} images, of which 4750 (95\%) normal images are from the class `4' and 250 (5\%) anomalous images are sampled randomly from the remaining nine classes. 

\noindent \textbf{Baseline Method} We use iForest \cite{liu2008} as the anomaly detector. We use the implementation in version 0.19.1 of the scikit-learn package \cite{scikit-learn} and vary the contamination hyperparameter from 0.01 to 0.69 in increments of 0.02 to generate the ROC curve for calculating AUC.


\noindent \textbf{Baseline Data Augmentation Techniques} We compare with two data augmentation techniques - adding random noise and elastic deformation \cite{simard2003, wong2016}, an optimized deformation for image augmentation. 

\begin{itemize}
\item \textbf{Addition of Random Noise} We randomly select a sample from the original dataset and then randomly pick 10\% of the pixels in a sample and set the value of each selected pixel to 1 if the original pixel value is less than 0.5 and 0 if the original pixel value is more than 0.5.
\item \textbf{Elastic Deformation} We randomly select a sample from the original dataset and apply the elastic deformation algorithm with optimal settings of $\sigma=4$ and $\alpha=34$, described by Simard \emph{et al.} \cite{simard2003}.
\end{itemize}


In each case, we generate 500 augmented or synthesized samples and add them to the original dataset. We then train the detection algorithm on the 5500 samples while testing on 5000 samples without augmentation. 


\noindent \textbf{Experiment Details} For DOPING, we follow the Unlabeled AAE implementation described in Section \ref{sec:implementation}, since we consider anomaly detection as a primarily unsupervised task.


\begin{equation}
	\mathcal{GG}(\mu, \alpha, \beta) = \frac{\beta}{2\alpha\Gamma(1/\beta)}e^{{-(|x-\mu|)/\alpha}^\beta}
    \label{eqn:gg}
\end{equation}

For comparison across different priors, we use the same architecture as the Unlabeled AAE described earlier but replace the Gaussian prior with a generalized Gaussian (Eqn. \ref{eqn:gg}) of different $\beta$ parameters, where $\beta=2$ gives the regular Gaussian distribution and as $\beta\rightarrow\infty$, the distribution approaches a uniform distribution \cite{nadarajah2005}. 
For this experiment, we vary $\beta$ from 0.5 to 3, keeping $\alpha$ and $\mu$ (other parameters in the generalized Gaussian function) consistent at 10.0 and 0.0 respectively. 
We also implement a version with a prior distribution of an 8-D multivariate Gaussian with isotropic covariance matrix of $10\mathbf{I}$.


We first conduct analysis of the latent space with magnitude-based sampling as described in Section \ref{sec:toy_expt_setup}. Subsequently, we show the results with the full DOPING method described in Section \ref{sec:implementation}.


\subsection{Optimal Sampling with Unlabeled AAE}

Here we validate our hypothesis on a high-dimensional dataset and show the effectiveness of the Unlabeled AAE, which allows our method to be unsupervised.

\begin{figure}
	\centering
	\begin{subfigure}[b]{6cm}
	\centering
	\includegraphics[width=6cm]{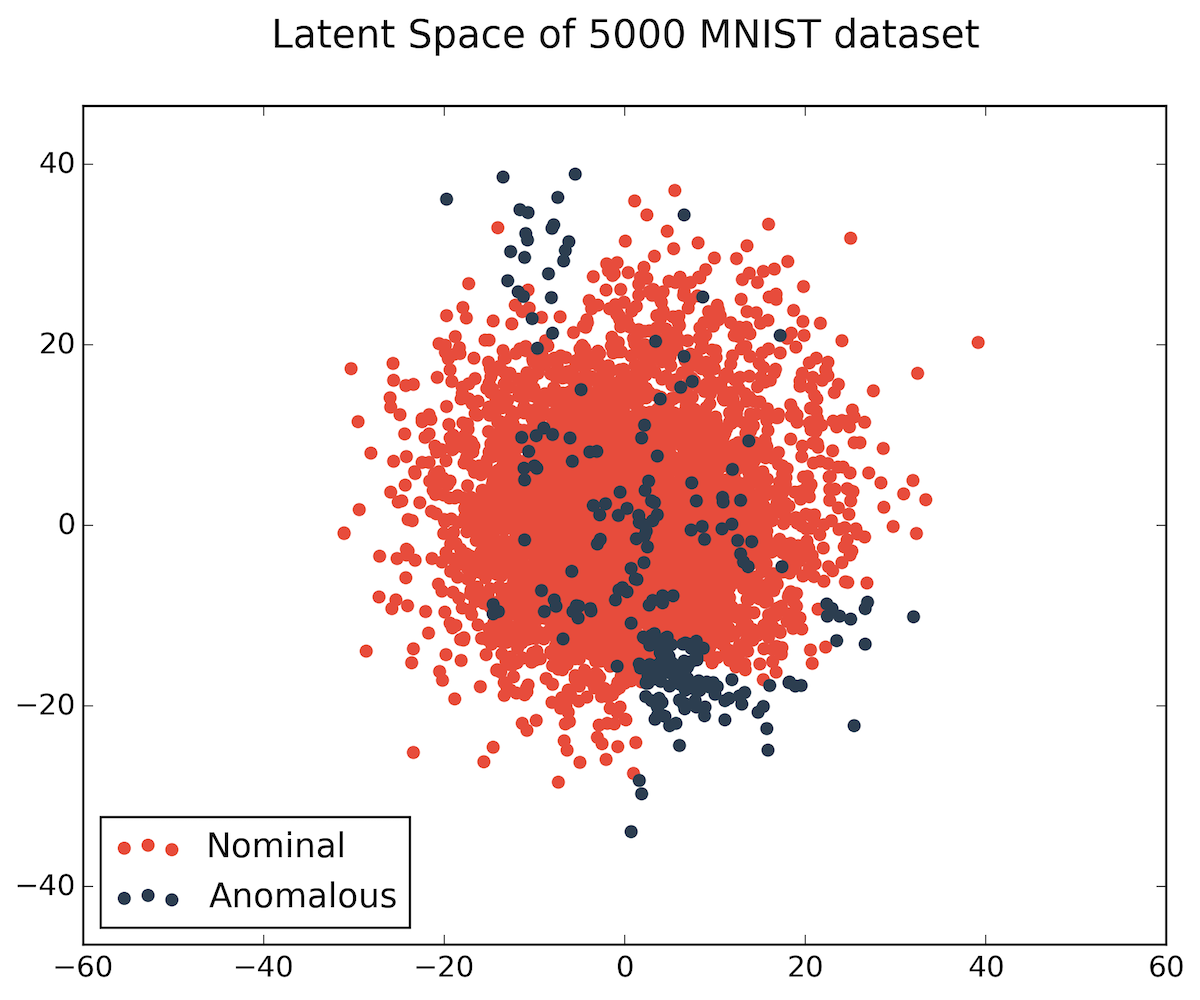}
	\caption{}
	\label{fig:mnist5000latent}
	\end{subfigure}
	\begin{subfigure}[b]{7cm}
	\centering
	\includegraphics[width=7cm]{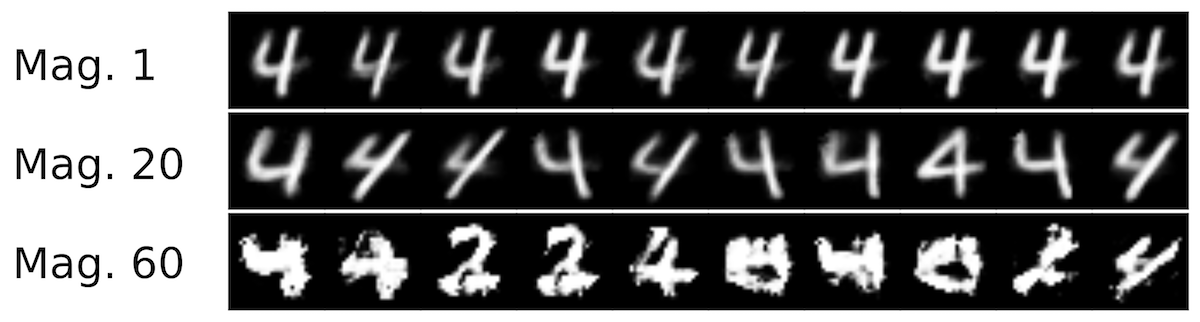}
	\caption{}
	\label{fig:mnist5000samples}
	\end{subfigure}
    \begin{subfigure}[b]{7cm}
	\centering
	\includegraphics[width=7cm]{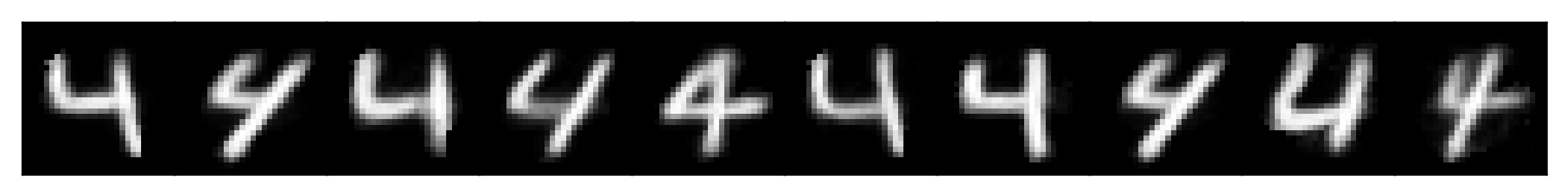}
	\caption{}
	\label{fig:mnist5000_edge}
	\end{subfigure}
	\caption{(a) 2D latent space encodings of the 5000 MNIST training images, (b) synthesized samples from latent vectors with $l^2$-norms/magnitudes 1 (inside), 20 (boundary) and 60 (outside) and (c) synthesized samples from latent vectors sampled with edge-based sampling.}
    \vspace{-0.5cm}
\end{figure}

Figure \ref{fig:mnist5000latent} shows the encodings of the original 5000 samples in the 2-D latent space, using an Unlabeled AAE, which follow a 2-D Gaussian. Since no label was used in the training of the AAE, both normal and anomalous samples are mixed in the distribution, although we observe some clustering of the anomalies.

Since the latent distribution follows a 2-D Gaussian, we are able to generate different probabilities of samples by sampling from rings of different radii, similar to the previous section. Figure \ref{fig:mnist5000samples} shows decoded samples drawn from different rings in the latent space. Intuitively, samples drawn close to the center of the latent distribution are more likely to look like common samples in the original data and hence more normal, whereas samples drawn from outside of the latent distribution show significant degradation. Finally, samples drawn from the boundary of the latent distribution resemble infrequent but normal images from the dataset.


\begin{figure}
	\centering
	\includegraphics[width=6cm]{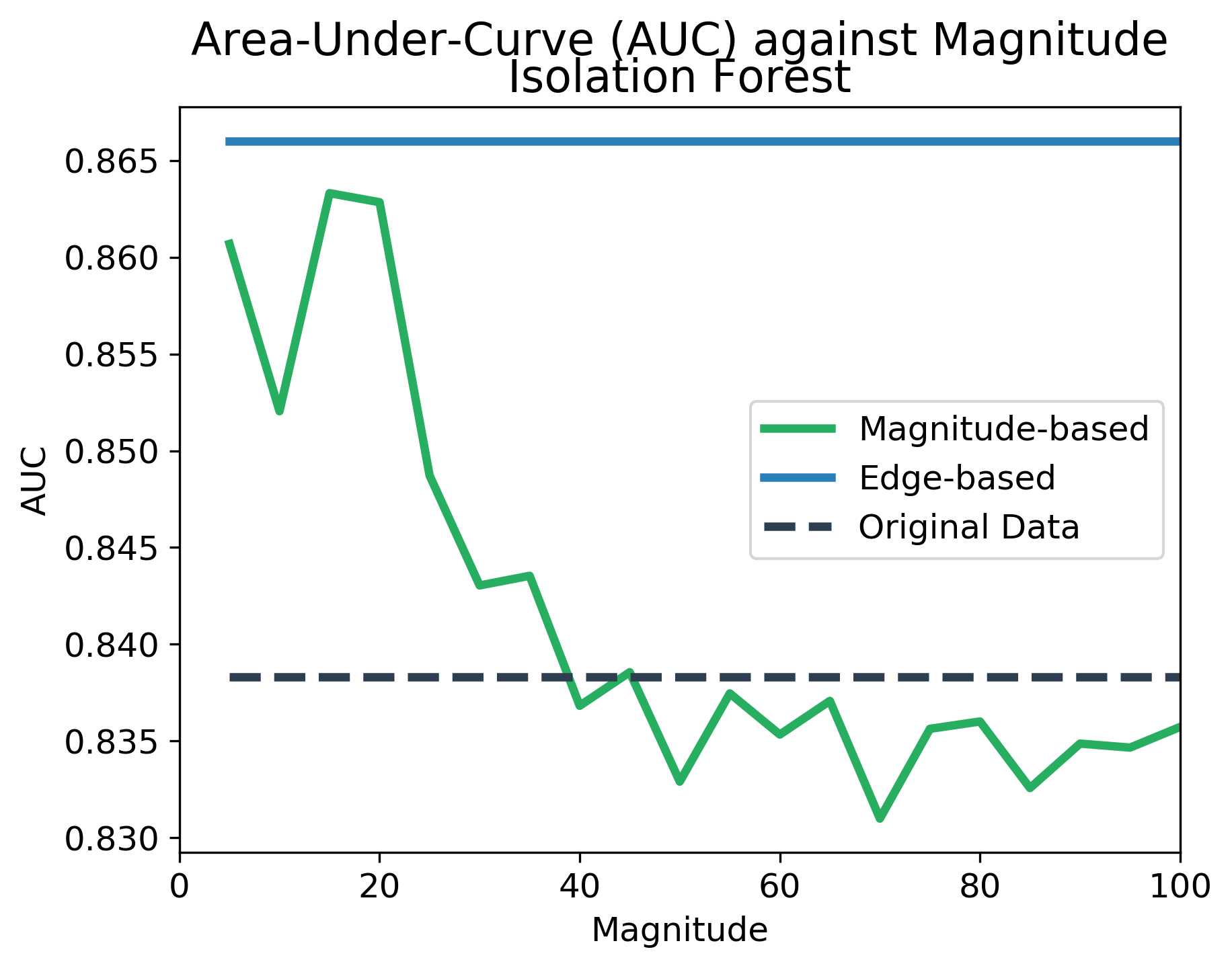}
	\caption{Graph of AUC against $l^2$-norm/magnitude of the latent vectors sampled for magnitude-based sampling applied to iForest on the MNIST experiment, compared against performance without DOPING and with edge-based DOPING.}
    \label{fig:mnist5000_iforest_auc}
    \vspace{-0.5cm}
\end{figure}

Figure \ref{fig:mnist5000_iforest_auc} plots AUC against $l^2$-norm/magnitude of sampled latent vectors for the iForest algorithm. We observe that the maximum AUC is achieved at magnitude 15.0, which again corresponds to the boundary of the latent distribution. 

Similar to the previous section, as we sample from increasing magnitudes, we see that the AUC worsens and falls below the performance of the original data without augmentation, implying that addition of samples at this region worsens the performance of the iForest algorithm. 

Finally, we also observe that addition of 500 samples generated using DOPING
reduces False Positive Rate from 0.31 to 0.26,
at a consistent True Positive Rate of 0.80.


\subsection{Comparison of Prior Distributions}
\label{sec:optimallatent}

Following the previous sections, 
we see that AAE is able to induce similar latent distributions 
for diverse datasets.
Through empirical results on the synthetic datasets, we also validate our hypothesis that decoding from samples at the boundary of the latent normal distribution results in the best improvement in anomaly detection and that we are able to do this with an Unlabeled AAE. In this section, we experiment with various prior distributions for generating artificial samples, using various settings from the generalized Gaussian \cite{nadarajah2005}, detailed in Section \ref{sec:mnist_setup}.


We use our method with the iForest algorithm for each of the prior distributions in Section \ref{sec:mnist_setup} and sample from a range of $l^2$-norms from the latent. Table \ref{tab:different_prior_results} shows the best AUC results obtained for each prior, which is consistently better than the performance of iForest on the original dataset without DOPING.

More interestingly, as $\beta$ increases, the optimal magnitude for obtaining the best AUC performance decreases. This makes sense intuitively, since the tails of the Generalized Gaussian grow lighter as $\beta$ increases, which also decreases the boundary of the distribution. In previous sections, we demonstrated that decoding from the boundary of the latent distribution gives significant improvement in anomaly detection performance. So, we see here that as $\beta$ increases, the boundary decreases and the optimal magnitude decreases as well.

This intuition is also true for the 8-D Gaussian. The distribution of $l^2$-norms for a multivariate Gaussian follows a $\chi$-distribution with $n$ degrees of freedom where $n$ is the number of dimensions for the original multivariate Gaussian. As $n$ increases, the mean of the $\chi$-distribution increases as well, which increases the boundary of the distribution. Hence we see here that the optimal magnitude for the 8-D Gaussian prior is higher than that of the 2-D Gaussian.


Here we show that the prior distribution in DOPING is not limited to the regular Gaussian ($\beta$ = 2). For the rest of the paper, we use the regular 2-D and 8-D Gaussian for simplicity.



\subsection{Comparison Against Other Data Augmentation Methods}
\label{sec:comparison}

Earlier, we demonstrated that the optimal region for sampling latent vectors is the edge of the latent distribution. From hereon, we show the results of the full DOPING method as in Algorithm \ref{alg:doping} and Section \ref{sec:implementation}. The edge-based sampling described in Section \ref{sec:implementation} allows us to reliably sample latent vectors from the edge of the distribution, without manually tuning the magnitude as in magnitude-based sampling.

Figure \ref{fig:mnist5000_edge} shows the decoded samples when we use edge-based sampling. Qualitatively, the edge-based samples are very similar to the samples obtained by magnitude-based sampling at magnitude 20, since the sampling from magnitude 20 also decodes from the boundary of the latent distribution.

We proceed to compare our method with other data augmentation methods when used with various state-of-the-art anomaly detection algorithms. For our method, we report results using the Unlabeled 2-D Gaussian prior described in the previous section. In addition, we also report a variant with an Unlabeled 8-D Gaussian prior, since \cite{makhzani2015} recommends a 5-D to 8-D latent space to adequately encode the MNIST dataset. 

\begin{table}
  \caption{AUC performance of DOPING with different priors, using the iForest algorithm in MNIST experiment.}
  \begin{tabular}{rcccc}
    \toprule
    &AUC&Optimal Mag.\\
     \midrule
     Original Dataset&0.838&-\\
     2-D Generalized Gaussian ($\beta=0.5$)&0.869&180.0\\
     2-D Generalized Gaussian ($\beta=1$)&0.870&25.0\\
     2-D Regular Gaussian ($\beta=2$)&0.863&15.0\\
     2-D Generalized Gaussian ($\beta=3$)&0.866&11.0\\
     8-D Regular Gaussian&0.874&41.0\\
   \bottomrule
 \end{tabular}
 \label{tab:different_prior_results}
 \end{table}

\begin{table}
\centering
  \caption{AUC Performance of the iForest algorithm with different data augmentations in the MNIST experiment.}
  \begin{tabular}{rcccc}
    \toprule
    &AUC\\
    \midrule
    No Augmentation&0.838\\
    Random Noise&0.842\\
    Elastic Deformation&0.843\\
    DOPING 2-D (Our Method)&0.866\\
    DOPING 8-D (Our Method)&\textbf{0.874}\\
  \bottomrule
\end{tabular}
\label{tab:mnist5000_results}
\vspace{-0.5cm}
\end{table}



\begin{figure}
	\centering
	\begin{subfigure}[b]{8cm}
	\centering
	\includegraphics[width=8cm]{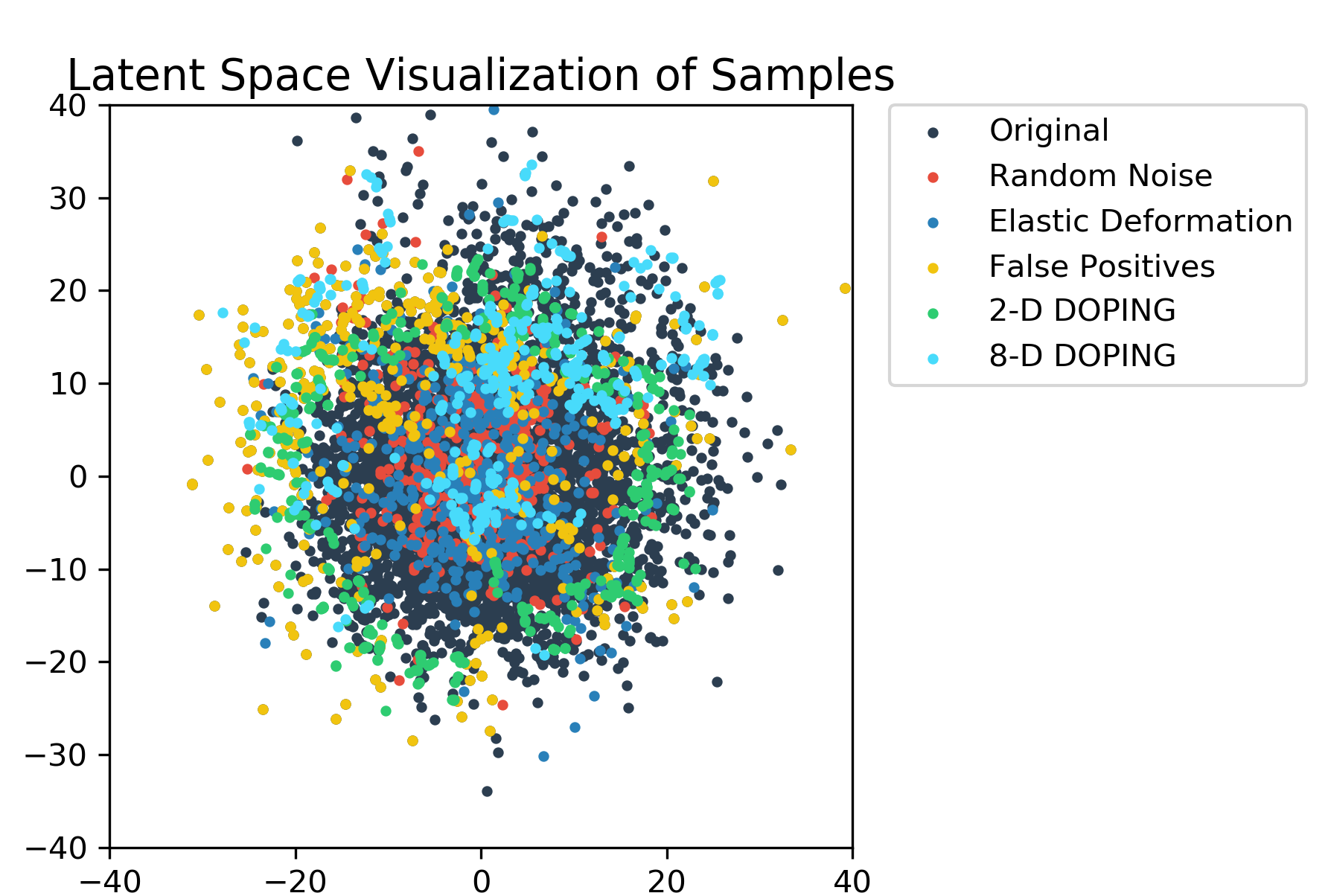}
	\caption{}
	\label{fig:mnist5000_augmented_encodings}
	\end{subfigure}
	\begin{subfigure}[b]{8cm}
	\centering
	\includegraphics[width=8cm]{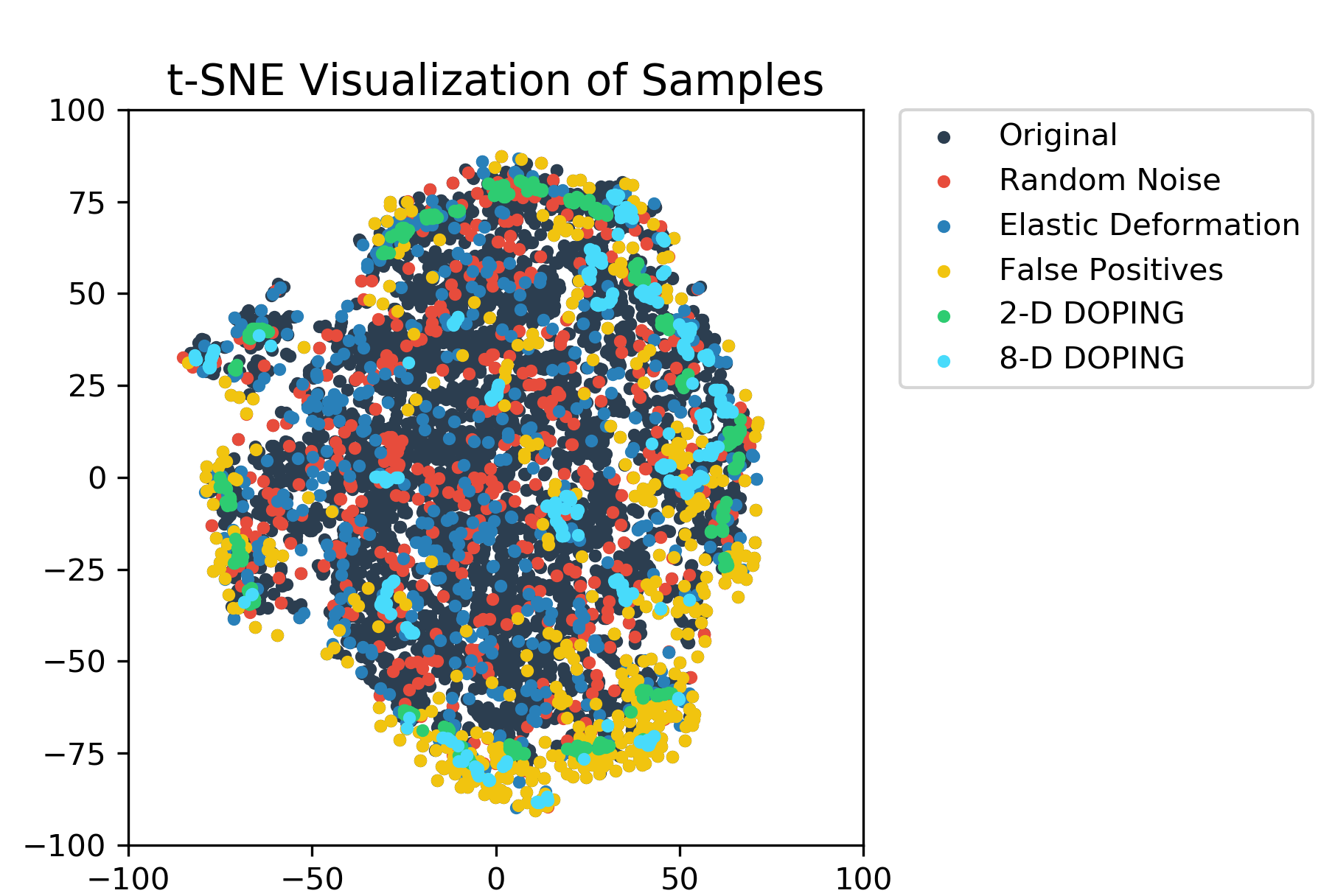}
	\caption{}
	\label{fig:mnist5000_augmented_encodings_tsne}
	\end{subfigure}
	\caption{(a) Latent space encodings of samples from augmentation methods in the MNIST experiment and (b) same samples visualized with t-SNE \cite{maaten2008}. Note that Common False Positives and DOPING samples occur in similar regions.}
\vspace{-0.5cm}
\end{figure}

\noindent \textbf{Results} Table \ref{tab:mnist5000_results} shows the AUC performance of Isolation Forest in the MNIST experiment, with different types of data augmentation. In general, we see that data augmentation methods provide moderate improvement to the AUC performance. However, both 2-D and 8-D DOPING consistently show comparable or much better performance, with the 8-D DOPING variant performing far better. This is likely due to the higher latent dimension which allows the latent space to better encode the dataset.

Figure \ref{fig:mnist5000_iforest_auc} also compares the performance of edge-based sampling to the AUC results for magnitude-based sampling at different magnitudes. We see that edge-based sampling shows similar performance to the best result obtained by magnitude-based sampling.

\noindent \textbf{Latent Space and t-SNE Visualization} For visualization and analysis, we begin by isolating the samples that are frequently classified as false positives by the detectors when no augmentation is used.
This is done by tagging all the false positive samples that occur at the iForest contamination hyperparameter of 0.13. We accumulate a total of 411 samples, which we hereon denote as False Positives.

We then encode the False Positives, as well as the encodings of samples from the data augmentation techniques, back to the latent space to visualize the respective latent distributions. Figure \ref{fig:mnist5000_augmented_encodings} shows the encodings of samples from the data augmentation techniques, with 500 samples from each technique, as well as the 411 false positive samples. 

For Random Noise and Elastic Deformation, the distributions follow the distribution of the original dataset. In contrast, samples generated by 2-D and 8-D 
DOPING resemble ring-like distributions in the latent space. Most importantly, the latent distributions of the 2-D and 8-D DOPING samples overlap strongly with that of the False Positives. The similar clustering between the False Positives and the samples generated by our method are even more apparent when we visualize the images with the t-SNE algorithm \cite{maaten2008}. Again, the False Positives and samples generated by both 2-D and 8-D DOPING are significantly denser towards the edge of the distribution. 

This demonstrates that the samples generated by DOPING help to increase the density of these False Positives in the training set, which allows the anomaly detection algorithm to better recognize these samples as normal. It is important to note that DOPING is able to achieve this without explicitly going through a round of anomaly detection testing to highlight the false positives.

\section{Evaluation with Real-world Datasets}
\label{sec:landsat_expt}

Here, we evaluate the performance of the formal DOPING algorithm, as described in Section \ref{sec:implementation} and Algorithm \ref{alg:doping}, on several real-world datasets.

\subsection{Dataset and Experimental Setup}


\noindent \textbf{Datasets} We show the results of DOPING on four anomaly detection datasets from the Outlier Detection DataSets (ODDS) repository\footnote{http://odds.cs.stonybrook.edu/}.

\begin{itemize}
	\item \textbf{Mammography} The Mammography dataset \cite{lichman2013} is obtained from the ODDS repository, with each sample having 6 features. The dataset comprises of 11183 samples, including 260 anomalies (2.3\%contamination).
    \item \textbf{Thyroid} The Thyroid dataset \cite{lichman2013} is obtained from the ODDS repository, with each sample having 6 features. The dataset comprises of 3772 samples, including 93 anomalies (2.5\% contamination). 
    \item \textbf{Lymphography} The Lymphography dataset \cite{lichman2013} is obtained from the ODDS repository, with each sample having 18 features. The dataset comprises of 148 samples, including 6 anomalies (4.0\% contamination).
    \item \textbf{Cardiotocography} The Cardiotocography dataset \cite{lichman2013} is obtained from the ODDS repository, with each sample having 21 features. The dataset comprises of 1831 samples, including 176 anomalies (9.6\% contamination).
    
\end{itemize}

For all datasets, we follow the settings in \cite{zhai2016, zong2018} with completely clean
training data: in each run, we randomly sample 50\% of the data for training with the remaining 50\% reserved for testing, and only data samples from the normal class are used for training models.


\begin{table}
\centering
  \caption{Comparing best F1 on real-world datasets.}
  \begin{tabular}{rccccc}
    \toprule
    \multirow{2}{*}{Best F1} & \multirow{2}{*}{iForest \cite{liu2008}} & iForest & iForest & iForest\\
    && w. DOPING & w. SMOTE& w. INOS \\
    \midrule
    Mammo.&0.304&0.345&0.319&\textbf{0.351}\\
    Cardio.&0.795&\textbf{0.812}&0.771&0.773\\
    Thyroid&0.744&\textbf{0.771}&0.758&0.730\\
    Lympho.&0.614&\textbf{0.720}&0.657&0.655\\
    Average&0.614&\textbf{0.662}&0.626&0.627\\
  \bottomrule
\end{tabular}
\label{tab:real_f1_results}
\end{table}

\begin{table}
\centering
  \caption{Comparing G-measure on real-world datasets.}
  \begin{tabular}{rccccc}
    \toprule
    \multirow{2}{*}{G-measure} & \multirow{2}{*}{iForest \cite{liu2008}} & iForest & iForest & iForest\\
    && w. DOPING & w. SMOTE& w. INOS \\
    \midrule
    Mammo.&0.318&0.345&0.323&\textbf{0.361}\\
    Cardio.&0.787&\textbf{0.807}&0.779&0.780\\
    Thyroid&0.754&\textbf{0.775}&0.753&0.737\\
    Lympho.&0.665&\textbf{0.750}&0.700&0.707\\
    Average&0.631&\textbf{0.669}&0.639&0.646\\
  \bottomrule
\end{tabular}
\label{tab:real_g_results}
\vspace{-0.5cm}
\end{table}

%

\noindent \textbf{Baseline Method} iForest \cite{liu2008} is used as the anomaly detector. We use the implementation in version 0.19.1 of the scikit-learn package \cite{scikit-learn} and vary the contamination hyperparameter from 0.01 to 0.69 in increments of 0.02 to calculate the AUC and search for the best F1. 

For each dataset, we train iForest with four variants: the original training set and three training sets augmented with DOPING, a SMOTE variant \cite{chawla2002} and an INOS variant \cite{cao2013} respectively. For each augmentation method, we synthesize an additional 10\% of the original training set size.


\noindent \textbf{Experiment Details} When implementing DOPING on these datasets, we first train an Unlabeled AAE on the training sets with a 2-D regular Gaussian prior with standard deviation 10.0 on both dimensions. We then synthesize samples using edge-based sampling as described in Section \ref{sec:implementation}. 

For comparison, we also show the results of iForest with two other data augmentation methods. Since the earlier data augmentation methods in Section \ref{sec:mnist_expt} do not transfer readily to non-image domains, we use variants of SMOTE \cite{chawla2002} and INOS \cite{cao2013} as comparisons. For both of the variants, we randomly select samples from the training set and apply interpolation in the data space. In the original SMOTE and INOS algorithms, only the minority class is oversampled in this manner. Here we randomly sample from the entire training set since we do not have labels in unsupervised anomaly detection.

For the SMOTE variant, we implement interpolation with nearest neighbor (Algorithm \ref{alg:internn}) on randomly selected samples across the whole training set. For INOS, we implement the technique using the OSTSC package\cite{dixon2017ostsc} and set the \textit{r} variable as 0.8 where 80\% of the synthetic samples are generated using ESPO and 20\% using ADASYN.


\subsection{Results and Discussion}
\label{sec:evaluation}





Tables \ref{tab:real_f1_results} and \ref{tab:real_g_results} show the performance of Isolation Forest \cite{liu2008} with and without augmentation, via F1 score and G-measure, where F1 is the harmonic mean of precision and recall and G-measure is the geometric mean.

We first observe that training with the augmented datasets mostly give better results than baseline iForest, implying that data augmentation has a positive effect on anomaly detection. Since a clean training set is used here, we postulate that general data augmentation of the uncontaminated training set helps to increase the density and variety of normal data samples in the training set.

Furthermore, we see that iForest with DOPING demonstrates the best performance across all datasets, with the exception of Mammography. Rather than randomly synthesizing training samples as with other augmentation methods, DOPING purposefully synthesizes infrequent normal samples and the better performance may be attributed to the anomaly detector better recognizing these infrequent normal samples in the dataset. 



\section{Conclusion and Future Work}

This paper proposes a novel form of data augmentation designed to tackle the problem where infrequent normal instances are misclassified, thereby reducing false positives. This is done by using an AAE to impose a multivariate Gaussian on the latent space and subsequently decoding from the edge of this latent distribution to generate infrequent normal samples.

In this paper we explained the intuition behind why DOPING helps to improve the performance of anomaly detectors and analyzed the performance of DOPING on several datasets. We show that the AUC against $l^2$-norm/magnitude trend is consistent across different datasets and DOPING makes use of this consistent trend to synthesize specific samples for dataset augmentation. Our experiments demonstrate empirically that our data augmentation technique helps to improve anomaly detection performance when applied across a variety of datasets.

Finally, the experiments reported here are meant to demonstrate the potential of our method and used a plain multivariate Gaussian prior with a fixed mean and covariance. More work can be done to investigate the effects of varying the distribution parameters and the type of distribution used and optimize the performance improvement on the unsupervised anomaly detection task.
In addition, the proposed method can be applied for some large-scale anomaly detection, e.g. network-wide anomalous traffic detection \cite{Manas:2018}.

\section*{Acknowledgment}
This work was supported by both ST Electronics and the National Research Foundation (NRF), Prime Minister's Office, Singapore under Corporate Laboratory @ University Scheme (Programme Title: STEE Infosec - SUTD Corporate Laboratory). The authors would also like to thank the anonymous reviewers for their valuable comments and helpful suggestions.

\bibliographystyle{IEEEtran}
\bibliography{main.bbl}

\begin{thebibliography}{10}
\providecommand{\url}[1]{#1}
\csname url@samestyle\endcsname
\providecommand{\newblock}{\relax}
\providecommand{\bibinfo}[2]{#2}
\providecommand{\BIBentrySTDinterwordspacing}{\spaceskip=0pt\relax}
\providecommand{\BIBentryALTinterwordstretchfactor}{4}
\providecommand{\BIBentryALTinterwordspacing}{\spaceskip=\fontdimen2\font plus
\BIBentryALTinterwordstretchfactor\fontdimen3\font minus
  \fontdimen4\font\relax}
\providecommand{\BIBforeignlanguage}[2]{{%
\expandafter\ifx\csname l@#1\endcsname\relax
\typeout{** WARNING: IEEEtran.bst: No hyphenation pattern has been}%
\typeout{** loaded for the language `#1'. Using the pattern for}%
\typeout{** the default language instead.}%
\else
\language=\csname l@#1\endcsname
\fi
#2}}
\providecommand{\BIBdecl}{\relax}
\BIBdecl

\bibitem{chawla2002}
N.~V. Chawla, K.~W. Bowyer, L.~O. Hall, and W.~P. Kegelmeyer, ``Smote:
  Synthetic minority over-sampling technique,'' \emph{Journal of artificial
  intelligence research}, vol.~16, pp. 321--357, 2002.

\bibitem{cao2011}
H.~Cao, X.-L. Li, Y.-K. Woon, and S.-K. Ng, ``Spo: Structure preserving
  oversampling for imbalanced time series classification,'' in \emph{Data
  Mining (ICDM), 2011 IEEE 11th International Conference on}.\hskip 1em plus
  0.5em minus 0.4em\relax IEEE, 2011, pp. 1008--1013.

\bibitem{chandola2009anomaly}
V.~Chandola, A.~Banerjee, and V.~Kumar, ``Anomaly detection: A survey,''
  \emph{ACM computing surveys (CSUR)}, vol.~41, no.~3, p.~15, 2009.

\bibitem{breunig2000}
M.~M. Breunig, H.-P. Kriegel, R.~T. Ng, and J.~Sander, ``Lof: Identifying
  density-based local outliers,'' in \emph{ACM sigmod record}, vol.~29,
  no.~2.\hskip 1em plus 0.5em minus 0.4em\relax ACM, 2000, pp. 93--104.

\bibitem{liu2008}
F.~T. Liu, K.~M. Ting, and Z.-H. Zhou, ``Isolation forest,'' in \emph{Data
  Mining, 2008. ICDM'08. Eighth IEEE International Conference on}.\hskip 1em
  plus 0.5em minus 0.4em\relax IEEE, 2008, pp. 413--422.

\bibitem{makhzani2015}
A.~Makhzani, J.~Shlens, N.~Jaitly, I.~Goodfellow, and B.~Frey, ``Adversarial
  autoencoders,'' \emph{arXiv preprint arXiv:1511.05644}, 2015.

\bibitem{goodfellow2014}
I.~Goodfellow, J.~Pouget-Abadie, M.~Mirza, B.~Xu, D.~Warde-Farley, S.~Ozair,
  A.~Courville, and Y.~Bengio, ``Generative adversarial nets,'' in
  \emph{Advances in neural information processing systems}, 2014, pp.
  2672--2680.

\bibitem{schlegl-ipmi-2017}
\BIBentryALTinterwordspacing
T.~Schlegl, P.~Seeb{\"{o}}ck, S.~M. Waldstein, U.~Schmidt{-}Erfurth, and
  G.~Langs, ``Unsupervised anomaly detection with generative adversarial
  networks to guide marker discovery,'' \emph{CoRR}, vol. abs/1703.05921, 2017.
  [Online]. Available: \url{http://arxiv.org/abs/1703.05921}
\BIBentrySTDinterwordspacing

\bibitem{ravanbakhsh-arxiv-2017}
M.~Ravanbakhsh, E.~Sangineto, M.~Nabi, and N.~Sebe, ``Training adversarial
  discriminators for cross-channel abnormal event detection in crowds,''
  \emph{CoRR}, vol. abs/1706.07680, 2017.

\bibitem{zenati2018}
H.~Zenati, C.~S. Foo, B.~Lecouat, G.~Manek, and V.~R. Chandrasekhar,
  ``Efficient gan-based anomaly detection,'' \emph{arXiv preprint
  arXiv:1802.06222}, 2018.

\bibitem{perez2017effectiveness}
L.~Perez and J.~Wang, ``The effectiveness of data augmentation in image
  classification using deep learning,'' \emph{arXiv preprint arXiv:1712.04621},
  2017.

\bibitem{antoniou2017}
A.~Antoniou, A.~Storkey, and H.~Edwards, ``Data augmentation generative
  adversarial networks,'' \emph{arXiv preprint arXiv:1711.04340}, 2017.

\bibitem{zhu2017}
\BIBentryALTinterwordspacing
X.~Zhu, Y.~Liu, Z.~Qin, and J.~Li, ``Data augmentation in emotion
  classification using generative adversarial networks,'' \emph{CoRR}, vol.
  abs/1711.00648, 2017. [Online]. Available:
  \url{http://arxiv.org/abs/1711.00648}
\BIBentrySTDinterwordspacing

\bibitem{sixt2016}
L.~Sixt, B.~Wild, and T.~Landgraf, ``Rendergan: Generating realistic labeled
  data,'' \emph{arXiv preprint arXiv:1611.01331}, 2016.

\bibitem{wong2016}
S.~C. Wong, A.~Gatt, V.~Stamatescu, and M.~D. McDonnell, ``Understanding data
  augmentation for classification: when to warp?'' in \emph{Digital Image
  Computing: Techniques and Applications (DICTA), 2016 International Conference
  on}.\hskip 1em plus 0.5em minus 0.4em\relax IEEE, 2016, pp. 1--6.

\bibitem{simard2003}
P.~Y. Simard, D.~Steinkraus, J.~C. Platt \emph{et~al.}, ``Best practices for
  convolutional neural networks applied to visual document analysis.'' in
  \emph{ICDAR}, vol.~3, 2003, pp. 958--962.

\bibitem{munawar-arxiv-2017}
A.~Munawar, P.~Vinayavekhin, and G.~D. Magistris, ``Limiting the reconstruction
  capability of generative neural network using negative learning,''
  \emph{CoRR}, vol. abs/1708.08985, 2017.

\bibitem{hasan-cvpr-2016}
M.~Hasan, J.~Choi, j.~Neumann, A.~K. Roy-Chowdhury, and L.~Davis, ``Learning
  temporal regularity in video sequences,'' in \emph{CVPR}, 2016.

\bibitem{zhou2017}
C.~Zhou and R.~C. Paffenroth, ``Anomaly detection with robust deep
  autoencoders,'' in \emph{Proceedings of the 23rd ACM SIGKDD International
  Conference on Knowledge Discovery and Data Mining}.\hskip 1em plus 0.5em
  minus 0.4em\relax ACM, 2017, pp. 665--674.

\bibitem{seebock-arxiv-2016}
P.~Seeb{\"{o}}ck, S.~M. Waldstein, S.~Klimscha, B.~S. Gerendas, R.~Donner,
  T.~Schlegl, U.~Schmidt{-}Erfurth, and G.~Langs, ``Identifying and
  categorizing anomalies in retinal imaging data,'' \emph{CoRR}, vol.
  abs/1612.00686, 2016.

\bibitem{xu2015learning}
D.~Xu, E.~Ricci, Y.~Yan, J.~Song, and N.~Sebe, ``Learning deep representations
  of appearance and motion for anomalous event detection,'' \emph{arXiv
  preprint arXiv:1510.01553}, 2015.

\bibitem{isola-cvpr-2017}
P.~Isola, J.-Y. Zhu, T.~Zhou, and A.~A. Efros, ``Image-to-image translation
  with conditional adversarial networks,'' \emph{CVPR}, 2017.

\bibitem{donahue2016}
J.~Donahue, P.~Kr{\"a}henb{\"u}hl, and T.~Darrell, ``Adversarial feature
  learning,'' \emph{arXiv preprint arXiv:1605.09782}, 2016.

\bibitem{han2005}
H.~Han, W.-Y. Wang, and B.-H. Mao, ``Borderline-smote: a new over-sampling
  method in imbalanced data sets learning,'' in \emph{International Conference
  on Intelligent Computing}.\hskip 1em plus 0.5em minus 0.4em\relax Springer,
  2005, pp. 878--887.

\bibitem{cao2013}
H.~Cao, X.-L. Li, D.~Y.-K. Woon, and S.-K. Ng, ``Integrated oversampling for
  imbalanced time series classification,'' \emph{IEEE Transactions on Knowledge
  and Data Engineering}, vol.~25, no.~12, pp. 2809--2822, 2013.

\bibitem{kingma2013}
D.~P. Kingma and M.~Welling, ``Auto-encoding variational bayes,'' \emph{arXiv
  preprint arXiv:1312.6114}, 2013.

\bibitem{kingma2014}
D.~Kingma and J.~Ba, ``Adam: A method for stochastic optimization,''
  \emph{arXiv preprint arXiv:1412.6980}, 2014.

\bibitem{trung:2018:distgan}
N.-T. Tran, T.-A. Bui, and N.-M. Cheung, ``Dist-gan: An improved gan using
  distance constraints,'' in \emph{ECCV}, 2018.

\bibitem{scikit-learn}
F.~Pedregosa, G.~Varoquaux, A.~Gramfort, V.~Michel, B.~Thirion, O.~Grisel,
  M.~Blondel, P.~Prettenhofer, R.~Weiss, V.~Dubourg, J.~Vanderplas, A.~Passos,
  D.~Cournapeau, M.~Brucher, M.~Perrot, and E.~Duchesnay, ``Scikit-learn:
  Machine learning in {P}ython,'' \emph{Journal of Machine Learning Research},
  vol.~12, pp. 2825--2830, 2011.

\bibitem{lecun1998}
Y.~LeCun, ``The mnist database of handwritten digits,'' \emph{http://yann.
  lecun. com/exdb/mnist/}, 1998.

\bibitem{nadarajah2005}
S.~Nadarajah, ``A generalized normal distribution,'' \emph{Journal of Applied
  Statistics}, vol.~32, no.~7, pp. 685--694, 2005.

\bibitem{maaten2008}
L.~v.~d. Maaten and G.~Hinton, ``Visualizing data using t-sne,'' \emph{Journal
  of machine learning research}, vol.~9, no. Nov, pp. 2579--2605, 2008.

\bibitem{lichman2013}
\BIBentryALTinterwordspacing
M.~Lichman, ``{UCI} machine learning repository,'' 2013. [Online]. Available:
  \url{http://archive.ics.uci.edu/ml}
\BIBentrySTDinterwordspacing

\bibitem{zhai2016}
S.~Zhai, Y.~Cheng, W.~Lu, and Z.~Zhang, ``Deep structured energy based models
  for anomaly detection,'' in \emph{International Conference on Machine
  Learning}, 2016, pp. 1100--1109.

\bibitem{zong2018}
B.~Zong, Q.~Song, M.~R. Min, W.~Cheng, C.~Lumezanu, D.~Cho, and H.~Chen, ``Deep
  autoencoding gaussian mixture model for unsupervised anomaly detection,'' in
  \emph{International Conference on Learning Representations}, 2018.

\bibitem{dixon2017ostsc}
M.~Dixon, D.~Klabjan, and L.~Wei, ``Ostsc: Over sampling for time series
  classification in r,'' 2017.

\bibitem{Manas:2018}
M.~Khatua, S.~Safavi, and N.-M. Cheung, ``Sparse laplacian component analysis
  for internet traffic anomalies detection,'' \emph{{IEEE} Transactions on
  Signal and Information Processing over Networks}, 2018.

\end{thebibliography}

\end{document}